\documentclass[conference]{IEEEtran}
\IEEEoverridecommandlockouts
\usepackage{cite}
\usepackage{amsmath,amssymb,amsfonts}
\usepackage{algorithmic}
\usepackage{graphicx}
\usepackage{textcomp}
\usepackage{xcolor}
\usepackage{enumitem} 
\usepackage{multirow}
\usepackage{booktabs} 
\usepackage{listings}  
\usepackage{adjustbox}
\usepackage{array}
\usepackage{geometry}
\usepackage{caption}
\usepackage{hyperref}
\usepackage{float}
\def\BibTeX{{\rm B\kern-.05em{\sc i\kern-.025em b}\kern-.08em
    T\kern-.1667em\lower.7ex\hbox{E}\kern-.125emX}}

\begin{document}

\title{HI-SQL: Optimizing Text-to-SQL Systems through Dynamic Hint Integration}


\author{\IEEEauthorblockN{Ganesh Parab\IEEEauthorrefmark{1}, 
Zishan Ahmad\IEEEauthorrefmark{2}, and
Dagnachew Birru\IEEEauthorrefmark{3}}
\IEEEauthorblockA{Quantiphi Analytics Solutions Pvt Ltd}\\
Email: \IEEEauthorrefmark{1}ganesh.parab@quantiphi.com,
\IEEEauthorrefmark{2}zishan.ahmad@quantiphi.com,
\IEEEauthorrefmark{3}dagnachew.birru@quantiphi.com
}

\maketitle

\begin{abstract}
Text-to-SQL generation bridges the gap between natural language and databases, enabling users to query data without requiring SQL expertise. While large language models (LLMs) have significantly advanced the field, challenges remain in handling complex queries that involve multi-table joins, nested conditions, and intricate operations. Existing methods often rely on multi-step pipelines that incur high computational costs, increase latency, and are prone to error propagation. 
To address these limitations, we propose HI-SQL, a pipeline that incorporates a novel hint generation mechanism utilizing historical query logs to guide SQL generation. By analyzing prior queries, our method generates contextual hints that focus on handling the complexities of multi-table and nested operations. These hints are seamlessly integrated into the SQL generation process, eliminating the need for costly multi-step approaches and reducing reliance on human-crafted prompts. Experimental evaluations on multiple benchmark datasets demonstrate that our approach significantly improves query accuracy of LLM-generated queries while ensuring efficiency in terms of LLM calls and latency, offering a robust and practical solution for enhancing Text-to-SQL systems.
\end{abstract}


\section{Introduction}
\label{se:intro}
The task of converting natural language queries into SQL statements, known as Text-to-SQL parsing, has consistently been a major area of interest within both the natural language processing and database communities. This technology enables users to interact with databases without requiring proficiency in SQL, thereby democratizing data access and facilitating informed decision-making across various domains \cite{b9}.

Early Text-to-SQL systems relied heavily on rule-based methods and human-engineered features, which, while effective to a degree, often struggled with scalability and adaptability to diverse database schemas \cite{b10}. The advent of deep learning and neural network models marked a significant shift, introducing sequence-to-sequence architectures that improved the translation of natural language to SQL \cite{b11}.

Current advancements in large language models (LLMs) have further propelled the field, enhancing the understanding of natural language semantics and the generation of accurate SQL queries. Recent studies have highlighted the impact of LLMs on Text-to-SQL parsing, noting their ability to capture complex linguistic patterns and improve performance across benchmarks \cite{b12}.

The development of large-scale human-labeled data sets, such as Spider \cite{b13} and BIRD \cite{b14}, has been instrumental in evaluating and advancing Text-to-SQL systems. These datasets provide diverse and complex queries that test the robustness and adaptability of current models.

Despite significant advancements, generating accurate SQL queries for complex operations that span multiple tables, nested conditions, and intricate joins continues to be a challenge within Text-to-SQL systems. Recent efforts have focused on leveraging LLMs to address these issues by designing agentic systems that combine multi-step processes with verification methods for generating SQL queries \cite{b8, b15, b16}. While these approaches have demonstrated effectiveness in improving query accuracy, they also lead to increased computational overhead due to the necessity of chaining multiple LLM calls.  This results in high token usage and latency, while also making the system vulnerable to error propagation, as inaccuracies in one stage can negatively impact subsequent steps, ultimately compromising the system’s reliability.

In this paper, we propose HI-SQL -- a system that leverages historical query logs to enhance the Text-to-SQL translation process. By systematically analyzing previously executed queries, our method generates contextual hints designed to address the complexities of handling multi-table queries, nested operations, and intricate conditions. Then, these hints are seamlessly integrated into the SQL generation pipeline, guiding the system toward producing more accurate and efficient queries. This targeted approach directly addresses one of the primary weaknesses of LLM-generated queries — handling complex database operations — while simplifying the overall system architecture and improving efficiency.

Unlike traditional agentic systems, such as CHASE-SQL \cite{b15}, which often require labor-intensive human annotations or expert-crafted few-shot examples to design effective prompts, our approach automates the hint generation process. By eliminating the need for human or domain-specific input in prompt design, our method significantly reduces manual effort and enhances scalability and adaptability across various database environments.

We rigorously evaluate our approach across multiple benchmarks, including SPIDER, BIRD, and ACME Insurance\cite{b29}, to assess its effectiveness in diverse scenarios. Experimental results reveal substantial improvements over baseline models, and 
results comparable to open-source state-of-the-art methods in the respective benchmarks. demonstrating the value of incorporating query-based hints for enhancing the performance of Text-to-SQL systems. Our solution achieves better query accuracy while reducing computational expenses by minimizing the number of LLM calls required, making it an efficient alternative to existing state-of-the-art methods. The main contributions of this work are two-fold \textit{viz}:
\textit{(i).} We introduce an automatic hint generation mechanism using historical query logs to improve SQL generation for complex queries; and \textit{(ii).} We present a SQL generation pipeline that reduces computational cost and latency by minimizing LLM calls while maintaining accuracy comparable to the state-of-the-art methods.

These contributions underscore the importance of leveraging historical data insights for developing more efficient, accurate, and scalable Text-to-SQL systems, offering promising directions for future research in natural language interfaces for databases.

\begin{figure*}
    \centering
    \includegraphics[width=\linewidth]{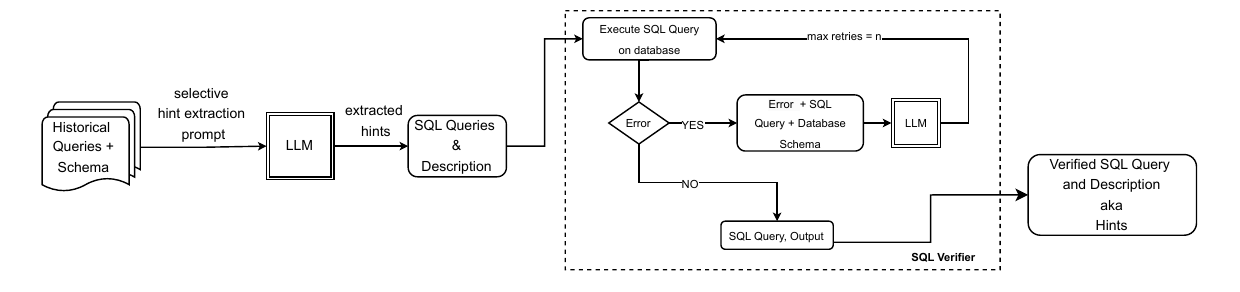}
    \caption{\textit{Hint Curation Pipeline:} The hint curation process utilizes historical queries and SQL Verifier to generate hints.}
    \label{fig:hintgen}
\end{figure*}

\begin{figure*}
    \centering
    \includegraphics[width=\linewidth]{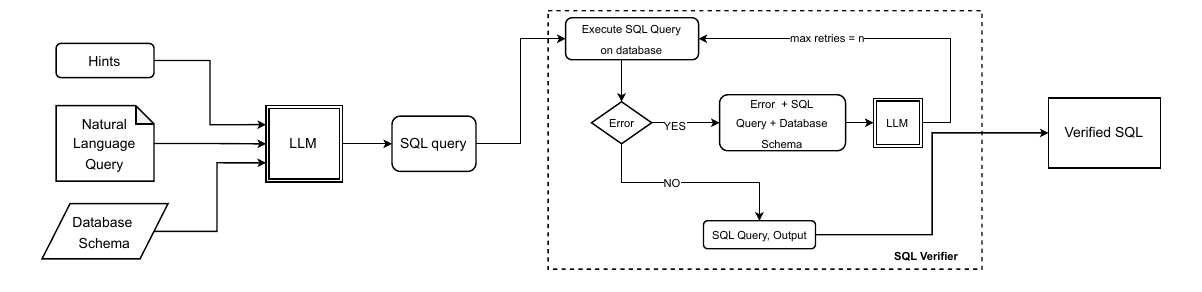}
    \caption{\textit{SQL Generation Pipeline:} This pipeline facilitates the SQL query generation process using previously generated hints and the SQL Verification process using the SQL Verifier pipeline.}
    \label{fig:sqlgen}
\end{figure*}

\section{Related Works}
\label{sec:related}
Text-to-SQL generation has evolved significantly, transitioning from early sequence-to-sequence models to advanced transformer-based architectures. Initial methods employed Recurrent Neural Networks (RNNs) and Long Short-Term Memory (LSTM) networks for encoding user queries and database schemas, utilizing slot-filling or auto-regressive decoding techniques to construct SQL queries. For instance, Seq2SQL \cite{b17} introduced reinforcement learning to enhance query generation, while SQLNet \cite{b18} employed a sketch-based approach to predict query structures before filling in details, thereby improving accuracy and efficiency. TypeSQL \cite{typesql} extended SQLNet by incorporating type information, enabling better handling of complex queries involving varied data types.

Advancements in the field of Text-to-SQL have led to the development of models like IRNet \cite{b19}, which introduced a graph-based encoder to capture relationships between database schema elements and user queries, improving performance on complex queries. RAT-SQL \cite{b20} further refined this approach by using a relation-aware transformer for schema encoding to manage multi-table queries. Meanwhile, SQLova \cite{b21} achieved significant progress by integrating BERT-style models to enhance schema representation and column attention, setting new benchmarks for Text-to-SQL systems.

Transformer-based models have played a pivotal role in advancing Text-to-SQL parsing. The T5 \cite{b22} model family, including T5-3B, demonstrated strong performance when fine-tuned on Text-to-SQL tasks. Domain-specific adaptations such as MedT5SQL \cite{b23} for healthcare and EDU-T5 for education showcased the potential of tailored models to address unique challenges in specialized fields. Additionally, PICARD \cite{b24} introduced constrained auto-regressive decoding, improving performance on dialogue-based and multi-turn SQL generation tasks.

The advent of large language models (LLMs) has redefined the field, with techniques like schema linking, self-correction, and self-consistency enhancing query accuracy for complex tasks. Benchmarks such as as Spider, BIRD, and other domain-specific datasets have facilitated the evaluation of these systems. More recently, specialized SQL agents have become one of the standard techniques for generating SQL queries. These agents often consist of multiple LLM calls for steps like entity disambiguation, schema linking, multi-SQL generation, self-correction/evolution etc \cite{b8, b26}. These agentic frameworks for SQL generation rely on multiple calls to LLMs and external tools, introducing latency and cost. Such frameworks often incorporate human-curated domain knowledge or few-shot prompts for enhanced performance. Another recent work \cite{b27} utilizes the human-curated ontology of the schema to generate cypher queries and then convert them back to SQL queries. 

Despite the current advancement, challenges remain in handling highly complex queries and improving efficiency. Our work addresses these gaps by leveraging historical query logs to generate contextual hints, enhancing accuracy, and reducing computational costs in Text-to-SQL systems.

\section{Methodology}
\label{sec:method}
Our pipeline consists of three modules, namely (i). Hint Curation, (ii). SQL Generation and (iii). SQL Verifier. In this section we describe each of these modules in detail.

\subsection{Hint Curation}
\label{sec:hint}
The  Hint Curation (c.f Figure \ref{fig:hintgen}) process leverages a large language model (LLM) to select and refine a diverse set of historical queries that reflect varying levels of complexity. These queries encompass tasks such as managing relationships between multiple tables, employing advanced data linking methods, applying conditional filtering, and performing computational operations like aggregations and averages. By relying on the LLM, this approach ensures a comprehensive exploration of diverse query structures and challenges commonly faced in database systems, minimizing the need for manual selection. Historical queries, along with their associated database schemas, are combined into structured prompts that guide the LLM in analyzing and identifying representative examples of complex query patterns. These prompts assist the LLM in extracting queries that showcase critical relationships between tables and address broader challenges in query design, including dynamic filtering, recursive structures, and integrating data from multiple sources. The extracted queries are run through the hosted database to verify their syntactical correctness. If any queries are found to be incorrect, the LLM automatically corrects them, thereby reducing reliance on manual review and correction. In addition to handling multi-table relationships and filtering, the LLM is tasked with recognizing more advanced complexities, such as hierarchical data structures, and analytical computations (e.g., row-based calculations), as well as set operations like unions, intersections, and differences. It identifies key patterns in table joins for a database schema, such as equi-joins, non-equi-joins, and self-joins, while also understanding how grouping and sorting enhance query outputs.

The final curated set of SQL queries (c.f Table \ref{tab:hints_example}), accompanied by detailed descriptions, provides actionable examples that improve the LLM’s ability to handle advanced query design. By automating this hint-generation process, the need for human-driven tasks such as creating ontologies or crafting few-shot examples for multi-agent systems-is significantly reduced. This ensures a faster and more scalable approach to enhancing the LLM's understanding of the complexities in query construction while also improving performance.

\subsection{SQL Generation}
\label{sec:sql_gen}
For  SQL generation (c.f Figure \ref{fig:sqlgen}), the LLM is provided with hints extracted through the Hint Curation process, along with the database schema and the Natural Language Query (NLQ), all formatted within a structured prompt. With the growing context window of modern LLMs, we do not utilise a query-specific schema filtering or linking module, thereby reducing the error incurred during schema linking \cite{b28}. This enables the LLM to generate an SQL query corresponding to the NLQ. The generated SQL is then processed through an SQL verifier to ensure its correctness. 

\subsection{ SQL Verification }
\label{sec: sql_verification}
The SQL verification pipeline ensures the correctness of the generated SQL by executing it on a hosted database and checking for both results and errors. If errors are encountered, the error message, along with the schema and the NLQ, is fed back into the LLM to generate a corrected SQL query. To address the non-deterministic and hallucinatory tendencies of LLMs, a retry loop has been implemented. If the LLM fails to correct the query, the verification process is retried up to a maximum of $C$ attempts. If all retries fail, the pipeline logs the error along with the NLQ for further manual intervention.

\begin{table*}[hbt!]
    \centering
    \caption{\label{tab:hints_example}A sample of hints for the financial database from BIRD, generated by our hint curation pipeline.}
    
    \begin{tabular}{p{0.95\textwidth}} 
        \begin{lstlisting}[basicstyle=\ttfamily\footnotesize, breaklines=true, caption=, captionpos=b, frame=single]

  {
        "description": "Join clients with their accounts and filter by specific account ID.",
        "sql_query": "SELECT STRFTIME('%Y', T1.birth_date) FROM client AS T1 INNER JOIN disp AS T3 ON T1.client_id = T3.client_id INNER JOIN account AS T2 ON T3.account_id = T2.account_id WHERE T2.account_id = 130"
    },
    {
        "description": "Join accounts with transactions to find the account opening date for a specific transaction amount and date.",
        "sql_query": "SELECT T1.date FROM account AS T1 INNER JOIN trans AS T2 ON T1.account_id = T2.account_id WHERE T2.amount = 840 AND T2.date = '1998-10-14'"
    },
    ...
    ...
    ...
    {
        "description": "Join districts with accounts to count how many accounts were opened in the branch with the largest number of crimes.",
        "sql_query": "SELECT COUNT(T2.account_id) FROM district AS T1 INNER JOIN account AS T2 ON T1.district_id = T2.district_id GROUP BY T1.A16 ORDER BY T1.A16 DESC LIMIT 1"
    }
        \end{lstlisting}
    \end{tabular}
\end{table*}

\begin{table*}[hbt!]
    \centering
    \caption{\label{tab:alldata}Performance of HI-SQL in comparison with the Baseline on BIRD, SPIDER-COMPLEX, and ACME-INSURANCE dataset}
   \begin{tabular}{llcccc}
   \toprule
\multirow{2}{*}{\textbf{Dataset}} & \multirow{2}{*}{\textbf{Method}} & \multicolumn{4}{c}{\textbf{Execution Accuracy}} \\ 
                                   &    & \textbf{Simple} & \textbf{Moderate} & \textbf{Challenging} & \textbf{Total}\\ 
    \midrule
    \multirow{2}{*}{BIRD} & Baseline  & 61\% & 38\% & 41\%  &54.1\%\\ 
    & \textbf{HI-SQL} & \textbf{69\%}& \textbf{52\%} & \textbf{56\%}\ & \textbf{62.38\%}\\  
    \midrule
    \multirow{2}{*}{SPIDER-COMPLEX} & Baseline  & - & - & - & 46\% \\  
    & \textbf{HI-SQL}  & - & - & -  & \textbf{63\%} \\ 
    \midrule
   \multirow{2}{*}{ACME-INSURANCE}  
   & Baseline &-&-&-&     22\% \\
   &KG-SQL  & - & - & - & 54.2\% \\  
    & \textbf{HI-SQL} & - & - & -  & \textbf{77.14\%} \\ 
    \bottomrule
\end{tabular}
\end{table*}

\begin{table*}[t]
    \centering
    \caption{\label{tab:sdsdata}Performance of HI-SQL in comparison with Baseline and CHESS on SDS sample dataset provided by CHESS}
   \begin{tabular}{llcccc}
   \toprule
\multirow{2}{*}{\textbf{Dataset}} & \multirow{2}{*}{\textbf{Method}} & \multicolumn{4}{c}{\textbf{Execution Accuracy}} \\ 
                                   &    & \textbf{Simple} & \textbf{Moderate} & \textbf{Challenging} & \textbf{Total}\\ 
    \midrule
    
    \multirow{3}{*}{BIRD\_SDS} & Baseline & 34\% & 57\% & 40\% & 48\% \\
    & CHESS  & 66\% & 41\% & 50\%  &55\%\\
    & \textbf{HI-SQL} & \textbf{68\%}& \textbf{54\%} & 50\%\ & \textbf{61.9\%}\\  
    \bottomrule
\end{tabular}
\end{table*}

\section{Experiments}
\label{sec:expt}

To test the efficacy of our method, we conduct the following experiments:
\begin{itemize}
    \item \textit{\textbf{Baseline:}} We conduct a baseline experiment without utilizing the generated hints. In this experiment, the NLQ and the entire schema of the corresponding database are passed to the LLM, and the output SQL query is obtained. Due to the large context window of modern LLMs we do not perform schema filtering and linking \cite{b29}.
    \item \textit{\textbf{HI-SQL:}} In this experiment we utilize the hints generated for the given database utilizing the hint curation module. For hint generation we utilize a small sample from the test data, treating these as historical queries for the hint generation process. These samples are excluded from the final test sample on which we test HI-SQL. For a given database, hint generation is a one-time process in the experiment. For SQL generation (c.f Figure \ref{fig:sqlgen}), we pass the NLQ, the full schema of the corresponding database, and the corresponding hints to the LLM. The resulting SQL is then processed through the SQL verification pipeline to make any necessary syntactic or schema-relevant corrections. 
    \item \textbf{\textit{CHESS:}} To compare our method against the current SOTA in the BIRD benchmark, we also run the CHESS \cite{b8} pipeline and compare its results with those of HI-SQL using an equivalent set from the BIRD benchmark. CHESS is recognized as the top-performing open-source pipeline in the BIRD benchmark and employs multiple LLM calls along with self-verification unit tests to handle complex SQL query generation.
    \item \textbf{\textit{KG-SQL:}} This state-of-the art method in ACME Insurance benchmark was proposed by Sequeda et al. \cite{b29}. This method requires an ontological representation of the database structure constituting of business logic. This ontology is constructed manually utilizing expertise in ontology creation and business usage of the database. Utilizing this ontology, their pipeline first generates a SPARQL query from the input NLQ through an LLM. This SPARQL query is then converted back to its SQL equivalent utilizing another manually created mapping file. This mapping file consists of mapping logic between nodes or edges in the ontology to the relevant data in the database. Since this method is dependent on human expert based ontology creation, we obtain results from this pipeline only on the ACME Insurance dataset, for which the ontology and mapping file are already provided.
    
    
\end{itemize}



\begin{table*}[!t]
\caption{\label{tab:sql_evaluation}Qualitative Evaluation of SQL queries from CHESS and HI-SQL}
    \begin{tabular}{@{}p{4cm}p{11cm}@{}}
        \toprule
        \textit{Natural Language Query} & Name the account numbers of female clients who are the oldest and have the lowest average salary? \\ \\
        \textit{CHESS generated SQL query} & \texttt{SELECT T3.account\_id FROM client AS T1 INNER JOIN disp AS T2 ON T1.client\_id = T2.client\_id INNER JOIN account AS T3 ON T2.account\_id = T3.account\_id WHERE T1.gender = 'F' AND T3.district\_id = (SELECT district\_id FROM district ORDER BY A11 ASC LIMIT 1) ORDER BY T1.birth\_date ASC LIMIT 1        ----- bird -----        financial} \\ \\
        \textit{HI-SQL generated SQL query} & \texttt{SELECT T3.account\_id FROM client AS T1 INNER JOIN district AS T2 ON T1.district\_id = T2.district\_id INNER JOIN account AS T3 ON T2.district\_id = T3.district\_id WHERE T1.gender = 'F' ORDER BY T1.birth\_date ASC, T2.A11 ASC LIMIT 1} \\ \\
        \textit{Comments} & The hints likely emphasized the need for district\_id to link client, district, and account. As a result, the HI-SQL generated query correctly joins client → district → account using district\_id, ensuring all required fields are accessible. \\ 
        \midrule
        
        \textit{Natural Language Query} & When was the set of cards with "Ancestor's Chosen" released? \\ \\
        \textit{CHESS generated SQL query} & \texttt{SELECT releaseDate FROM sets WHERE code = (SELECT T1.setCode FROM set\_translations AS T1 INNER JOIN cards AS T2 ON T1.setCode = T2.setCode WHERE T2.name = 'Ancestor''s Chosen')        ----- bird -----        card\_games} \\ \\
        \textit{HI-SQL generated SQL query} & \texttt{SELECT T2.releaseDate FROM cards AS T1 INNER JOIN sets AS T2 ON T1.setCode = T2.code WHERE T1.name = 'Ancestor''s Chosen'} \\ \\
        \textit{Comments} & Hints likely emphasized simplifying the query by avoiding subqueries where direct joins can achieve the same result. The second query directly retrieves the releaseDate by joining cards with sets. \\ 
        \midrule
        
        \textit{Natural Language Query} & Please list the full names of all the superheroes with over 15 superpowers. \\ \\
        \textit{CHESS generated SQL query} & \texttt{SELECT DISTINCT T2.full\_name FROM superhero AS T2 INNER JOIN hero\_power AS T1 ON T2.id = T1.hero\_id GROUP BY T2.id HAVING COUNT(T1.power\_id) > 15        ----- bird -----        superhero} \\ \\
        \textit{HI-SQL generated SQL query} & \texttt{SELECT T1.full\_name FROM superhero AS T1 INNER JOIN hero\_power AS T2 ON T1.id = T2.hero\_id GROUP BY T1.full\_name HAVING COUNT(T2.power\_id) > 15} \\ \\
        \textit{Comments} & CHESS's query could create confusion because T2.full\_name is selected while grouping by T2.id. Without explicitly grouping by full\_name, the query relies on implicit relationships. HI-SQL's query explicitly groups by T1.full\_name, making the logic clear and unambiguous. CHESS's query groups by T2.id while selecting T2.full\_name, relying on implicit rules. HI-SQL generated query groups by T1.full\_name explicitly for correctness. \\ 
        \midrule
        
        \textit{Natural Language Query} & Which patient is the first patient with an abnormal anti-SSA to come to the hospital? Please give his or her ID. \\ \\
        \textit{CHESS generated SQL query} & \texttt{SELECT P.ID FROM Patient AS P INNER JOIN Laboratory AS L ON P.ID = L.ID WHERE L.SSA NOT IN ('-', '+-') ORDER BY P.`First Date` ASC LIMIT 1        ----- bird -----        thrombosis\_prediction} \\ \\
        \textit{HI-SQL generated SQL query} & \texttt{SELECT T1.ID FROM Patient AS T1 INNER JOIN Laboratory AS T2 ON T1.ID = T2.ID WHERE T1.`First Date` IS NOT NULL AND T2.SSA NOT IN ('negative', '0') ORDER BY T1.`First Date` ASC LIMIT 1} \\ \\
        \textit{Comments} & CHESS's query relies on symbols ('-' and '+-'), which might not adequately represent the intended "negative" or neutral values. If the SSA column uses descriptive labels (e.g., 'negative' or '0'), this condition may fail to filter correctly. HI-SQL's hints likely clarified that the SSA column should be filtered using more descriptive values such as 'negative'. The query adjusts the SSA filtering to use values aligned with standard descriptive labels, ensuring proper filtering. \\ 
        \bottomrule
    \end{tabular}
\end{table*}

\subsection{Datasets and Evaluation Metrics}
\begin{itemize}
    \item \textit{\textbf{BIRD:}} The BIRD dataset's development set comprises 1,534 samples, each pairing natural language questions with corresponding SQL queries. This set spans various databases, including domains such as healthcare, finance, and education, among others. The SQL queries in the BIRD dataset range from simple to highly complex, requiring models to handle intricate query structures such as advanced joins, recursive patterns, and multi-step aggregations. This makes BIRD a demanding benchmark for evaluating and enhancing text-to-SQL model capabilities.
    \item \textbf{\textit{Subsampled Development Set (SDS):}} To compare our method with another open-source SOTA method, we utilize a subsample of the BIRD dataset SDS provided by CHESS \cite{b8}. Due to the high number of LLM calls inherent in the CHESS pipeline, which results in significant costs, we choose to compare our method against this subsample and not on the entire BIRD dataset. This curated subset consists of 147 samples, categorized into three levels of difficulty: (i). Simple having 81 samples, (ii). Moderate, having 54 samples, and (iii). Challenging having 12 queries. The use of this SDS allows for an efficient evaluation across a range of query complexities.
    \item \textit{\textbf{SPIDER-COMPLEX:}} The SPIDER dataset is a large-scale, human-annotated resource for complex, cross-domain text-to-SQL tasks. It comprises over 10,000 natural language questions and 5,600 unique SQL queries across 200 databases spanning 138 domains. Unlike older datasets, SPIDER requires models to generalize to new SQL queries and database schemas not seen during training, making it a standard benchmark for evaluating text-to-SQL models. However, most queries in the dataset are not complex enough to require joins, nested conditions, or other multi-table operations. Since our pipeline focuses on solving complex queries, we filter out only those complex queries from the SPIDER dataset that involves multi-table operations. In total we identify 84 such queries within the entire SPIDER dataset for our experiments.
    \item \textbf{\textit{ACME-INSURANCE:}}
    This benchmark is built upon a subset of the Property and Casualty (P\&C) Data Model, developed by the Object Management Group (OMG) to meet the specific data management needs of the insurance industry. The complete physical schema comprises 199 tables, each with a primary key, and 243 foreign key relationships. Sequeda, et al. \cite{b29} released a reduced schema consisting of 13 tables.
    
    The benchmark introduced in their work is characterized by its significant complexity, which stems from the intricate design of its natural language queries. These queries are carefully designed to simulate real-world business problems, requiring the retrieval and integration of data from multiple interconnected tables within an enterprise SQL schema. The schema, tailored to the insurance domain, includes 13 interconnected tables representing entities such as policies, claims, and coverages. This structure reflects the highly interconnected nature of enterprise data. Effectively addressing the benchmark’s complex queries demands both syntactic accuracy and a deep semantic understanding of the data’s organization and the business logic it encapsulates.
\end{itemize}
\textit{\textbf{Evaluation Metric:}} We use Execution Accuracy (EA) as the metric to evaluate the performance of our method. Execution Accuracy measures the correctness of the SQL outputs by comparing the results of the predicted queries against those from ground-truth queries when executed on respective database instances. This metric provides a comprehensive assessment by accounting for variations in valid SQL queries that can yield identical results for a given question.

\subsection{Experimental Setup}

For the hint curation, SQL query generation, and SQL verification processes, we utilize OpenAI’s GPT-4o with a temperature setting of 0.3 to control the randomness of the generated output. 
The hint generation process involves a single call to the LLM to extract relevant hints for a specific database. 
The verification pipeline includes a feedback loop with a maximum of 3 retries ($C = 3$). This retry mechanism ensures that if the query is consistently re-generated incorrectly, the process terminates after $C$ attempts. For hint generation purposes, we used 20\% of samples from each dataset while excluding these samples from test sets when computing Execution Accuracy. All the experiments were evaluated against the 80\% held-out samples in each dataset.


\section{Results and Analysis}

To thoroughly evaluate the effectiveness of our proposed method, HI-SQL, we conduct a series of experiments using four datasets: BIRD, SPIDER-COMPLEX, ACME-INSURANCE, and BIRD SDS. These experiments benchmark HI-SQL against baseline, CHESS, and KG-SQL. The primary metric for evaluation is Execution Accuracy (EA), with additional focus on the number of LLM calls required for a complete operation.

The number of LLM calls in CHESS is $M \times K \times N$; where $M$ represents the number of unit tests, $K$ is the number of queries generated per turn, and $N$ is the total number of natural language queries. In contrast, our system is more efficient. We generate just one LLM call for hint generation across the entire database. If any query is invalid, the SQL verifier triggers one additional LLM call to rewrite the query. Additionally, our system requires one LLM call for each user-provided natural language query. The total number of LLM calls in our system is  $(C \times N)+(1 \times S)$, where $C$ is the number of self-verifications, $N$ is the total number of
natural language queries, and $S$ is the number of SQL re-generations (if required) for hints.


The experimental results underscore the superior performance of HI-SQL across diverse datasets. As summarized in Table \ref{tab:alldata}, HI-SQL achieves substantial improvements in execution accuracy compared to the baseline. On the BIRD dataset, HI-SQL attains a total execution accuracy of 62.38\%, outperforming the baseline's 54.1\%. When analyzing query complexities, HI-SQL exhibits gains of 8\%, 14\%, and 15\% in Simple, Moderate, and Challenging categories, respectively. Similar trends are observed for SPIDER-COMPLEX and ACME-INSURANCE datasets, where HI-SQL achieves total execution accuracies of 63\% and 77.14\%, surpassing the baseline of 17\% and 55\%, respectively.  Comparison results demonstrate that HI-SQL outperforms KG-SQL, achieving a higher accuracy of 77.14\% compared to 54.2\%. These results highlight HI-SQL’s ability to handle varying query complexities while maintaining high accuracy.

Furthermore, as shown in Table \ref{tab:sdsdata}, HI-SQL was evaluated against CHESS on the BIRD\_SDS dataset. Here, HI-SQL achieves a total execution accuracy of 61.9\%, surpassing CHESS's performance of 55\%. Notably, HI-SQL demonstrates a significant improvement of 13\% in Moderate complexity queries, showcasing its robustness in handling intermediate-level challenges.

The analysis of SQL query corrections reveals consistent patterns that enhance efficiency, accuracy, and clarity in query design. Common improvements include replacing subqueries with joins, which streamlines execution by reducing redundancy and introducing explicit aggregation and sorting to ensure clarity in computation and ordering logic. Queries were further simplified by removing unnecessary columns and computations, improving readability and performance. 
Semantic alignment was achieved by refining filtering conditions to match dataset characteristics, while null checks were added to prevent errors from missing values. Clear and consistent aliasing improved query readability, and string operations were optimized to reduce computational overhead. By analyzing historical query patterns, relationships, and data structures, hints provide the model with contextual guidance that aligns with the schema and query intent. These hints reduce ambiguity in generated queries, enabling the model to produce SQL that adheres to schema constraints, optimizes joins, and applies accurate filtering logic.

\section{Conclusion}

The experimental evaluation underscores the efficacy and efficiency of HI-SQL in generating accurate and optimized SQL queries across diverse datasets and query complexities. By leveraging a streamlined feedback loop and reducing the number of LLM calls, HI-SQL achieves significant computational efficiency compared to CHESS and other baseline methods. Its superior execution accuracy, with notable gains over baseline, CHESS, and KG-SQL across their specialized benchmarks, demonstrates its robustness in handling challenging scenarios.

Additionally, HI-SQL’s intelligent hint curation and correction mechanisms ensure semantic alignment with database schemas, enhancing query clarity, performance, and execution accuracy. This approach not only reduces resource consumption but also positions HI-SQL as a cost-effective solution for practical deployment in real-world SQL generation tasks. Unlike current systems that, once deployed, have no scope for improvement, our pipeline can utilize historical query logs to come up with new hints and thus adapt to give higher performance.



\begin{thebibliography}{00}

\bibitem{b8} Talaei, S., Pourreza, M., Chang, Y. C., Mirhoseini, A., \& Saberi, A. ``Chess: Contextual harnessing for efficient sql synthesis. '' arXiv preprint arXiv:2405.16755, 2024.
\bibitem{b9} Deng, N., Chen, Y., \& Zhang, Y. ``Recent advances in text-to-SQL: a survey of what we have and what we expect'', arXiv preprint arXiv:2208.10099, 2022.
\bibitem{b10} J. M. Zelle and R. J. Mooney, ``Learning to parse
database queries using inductive logic programming,'' in Proceedings of the national conference on artificial intelligence, 1996.
\bibitem{b11} I. Sutskever, O. Vinyals, and Q. V. Le, ``Sequence
to sequence learning with neural networks,'' Proc. of NeurIPS, 2014.
\bibitem{b12} Shi L, Tang Z, Zhang N, Zhang X, Yang Z. ``A survey on employing large language models for text-to-sql tasks'', arXiv preprint arXiv:2407.15186, 2024.
\bibitem{b13} Yu T, Zhang R, Yang K, Yasunaga M, Wang D, Li Z, Ma J, Li I, Yao Q, Roman S, Zhang Z. ``Spider: A large-scale human-labeled dataset for complex and cross-domain semantic parsing and text-to-sql task'', arXiv preprint arXiv:1809.08887, 2018.
\bibitem{b14} Li J, Hui B, Qu G, Yang J, Li B, Li B, Wang B, Qin B, Geng R, Huo N, Zhou X, ``Can llm already serve as a database interface? a big bench for large-scale database grounded text-to-sqls'', Advances in Neural Information Processing Systems, 2024.
\bibitem{b15} Gao Y, Liu Y, Li X, Shi X, Zhu Y, Wang Y, Li S, Li W, Hong Y, Luo Z, Gao J, ``XiYan-SQL: A Multi-Generator Ensemble Framework for Text-to-SQL'', arXiv preprint arXiv:2411.08599, 2024.
\bibitem{b16} Pourreza M, Li H, Sun R, Chung Y, Talaei S, Kakkar GT, Gan Y, Saberi A, Ozcan F, Arik SO, ``Chase-sql: Multi-path reasoning and preference optimized candidate selection in text-to-sql'', arXiv preprint arXiv:2410.01943, 2024.
\bibitem{b17} Zhong V, Xiong C, Socher R. ``Seq2sql: Generating structured queries from natural language using reinforcement learning'', arXiv preprint arXiv:1709.00103, 2017.
\bibitem{typesql} Yu T, Li Z, Zhang Z, Zhang R, Radev D, ``TypeSQL: Knowledge-Based Type-Aware Neural Text-to-SQL Generation'', InProceedings of the 2018 Conference of the North American Chapter of the Association for Computational Linguistics: Human Language Technologies, 2018.
\bibitem{b18} Xu X, Liu C, Song D, ``Sqlnet: Generating structured queries from natural language without reinforcement learning'', arXiv preprint arXiv:1711.04436, 2017.
\bibitem{b19} Guo J, Zhan Z, Gao Y, Xiao Y, Lou JG, Liu T, Zhang D, ``Towards complex text-to-sql in cross-domain database with intermediate representation'', arXiv preprint arXiv:1905.08205, 2019.
\bibitem{b20} Wang B, Shin R, Liu X, Polozov O, Richardson M, ``Rat-sql: Relation-aware schema encoding and linking for text-to-sql parsers'', arXiv preprint arXiv:1911.04942. November 2019.
\bibitem{b21} Hwang W, Yim J, Park S, Seo M, ``A comprehensive exploration on wikisql with table-aware word contextualization'', arXiv preprint arXiv:1902.01069. February 2019.
\bibitem{b22} Raffel C, Shazeer N, Roberts A, Lee K, Narang S, Matena M, Zhou Y, Li W, Liu PJ, ``Exploring the limits of transfer learning with a unified text-to-text transformer'', Journal of machine learning research, 2020, pp. 1--67.
\bibitem{b23} Marshan A, Almutairi AN, Ioannou A, Bell D, Monaghan A, Arzoky M. ``MedT5SQL: a transformers-based large language model for text-to-SQL conversion in the healthcare domain'', Frontiers in Big Data, June 2024.
\bibitem{b24} Scholak T, Schucher N, Bahdanau D, ``PICARD: Parsing incrementally for constrained auto-regressive decoding from language models'', arXiv preprint arXiv:2109.05093, September 2021.
\bibitem{b25} Qi J, Tang J, He Z, Wan X, Cheng Y, Zhou C, Wang X, Zhang Q, Lin Z, ``Rasat: Integrating relational structures into pretrained seq2seq model for text-to-sql'', arXiv preprint arXiv:2205.06983, May 2022.
\bibitem{b26} Pourreza M, Li H, Sun R, Chung Y, Talaei S, Kakkar GT, Gan Y, Saberi A, Ozcan F, Arik SO, ``Chase-sql: Multi-path reasoning and preference optimized candidate selection in text-to-sql'', arXiv preprint arXiv:2410.01943, October 2024.
\bibitem{b27} Caferoğlu HA, Ulusoy Ö, ``E-sql: Direct schema linking via question enrichment in text-to-sql'', arXiv preprint arXiv:2409.16751, September 2024.
\bibitem{b28} Maamari K, Abubaker F, Jaroslawicz D, Mhedhbi A, ``The death of schema linking? text-to-sql in the age of well-reasoned language models'', arXiv preprint arXiv:2408.07702, August 2024.
\bibitem{b29} Sequeda J, Allemang D, Jacob B, ``A benchmark to understand the role of knowledge graphs on large language model's accuracy for question answering on enterprise SQL databases'', InProceedings of the 7th Joint Workshop on Graph Data Management Experiences \& Systems (GRADES) an Network Data Analytics (NDA), June 2024, pp. 1--12.


\end{thebibliography}
\end{document}